\documentclass{article}

\usepackage{microtype}
\usepackage{graphicx}
\usepackage{subcaption}
\usepackage{booktabs} 

\usepackage{hyperref}

\usepackage{enumitem}

\usepackage{multirow}
\usepackage{pifont}
\usepackage[table]{xcolor}
\usepackage{pgf}
\usepackage{colortbl}
\usepackage{xcolor}

\usepackage{colortbl}
\usepackage{pgf} 

\newcommand{\method}{ProRes}

\usepackage[preprint]{icml2026}

\usepackage{amsmath}
\usepackage{amssymb}
\usepackage{mathtools}
\usepackage{amsthm}

\usepackage[capitalize,noabbrev]{cleveref}

\theoremstyle{plain}

\theoremstyle{definition}

\theoremstyle{remark}

\usepackage[disable,textsize=tiny]{todonotes}

\icmltitlerunning{Progressive Residual Warmup for Language Model Pretraining}

\begin{document}

\twocolumn[
  \icmltitle{Progressive Residual Warmup for Language Model Pretraining}



  \icmlsetsymbol{equal}{*}



  \begin{icmlauthorlist}
    \icmlauthor{Tianhao Chen}{hkust}
    \icmlauthor{Xin Xu}{hkust}
    \icmlauthor{Lu Yin}{surrey}
    \icmlauthor{Hao Chen}{hkust}
    \icmlauthor{Yang Wang}{hku}
    \icmlauthor{Shizhe Diao}{nvidia}
    \icmlauthor{Can Yang}{hkust}
  \end{icmlauthorlist}

  \icmlaffiliation{hkust}{The Hong Kong University of Science and Technology, Hong Kong, China}
  \icmlaffiliation{nvidia}{NVIDIA, Santa Clara, US}
  \icmlaffiliation{hku}{The University of Hong Kong, Hong Kong, China}
  \icmlaffiliation{surrey}{University of Surrey, Guildford, UK}


  \icmlcorrespondingauthor{Can Yang}{macyang@ust.hk}

  \icmlkeywords{Machine Learning, ICML}

  \vskip 0.3in
]



\printAffiliationsAndNotice{}  

\begin{abstract}
    Transformer architectures serve as the backbone for most modern Large Language Models, therefore their pretraining stability and convergence speed are of central concern. Motivated by the logical dependency of sequentially stacked layers, we propose \textbf{\underline{Pro}}gressive \textbf{\underline{Res}}idual Warmup (\textbf{\method{}}) for language model pretraining. \method{} implements an ``early layer learns first'' philosophy by multiplying each layer's residual with a scalar that gradually warms up from 0 to 1, with deeper layers taking longer warmup steps. In this way, deeper layers wait for early layers to settle into a more stable regime before contributing to learning. We demonstrate the effectiveness of \method{} through pretraining experiments across various model scales, as well as normalization and initialization schemes. Comprehensive analysis shows that \method{} not only stabilizes pretraining but also introduces a unique optimization trajectory, leading to faster convergence, stronger generalization and better downstream performance. Our code is available at \href{https://github.com/dandingsky/ProRes}{https://github.com/dandingsky/ProRes}.
\end{abstract}

\section{Introduction}

\begin{figure}[ht]
  \vskip 0.2in
  \begin{center}
    \centerline{\includegraphics[width=\columnwidth]{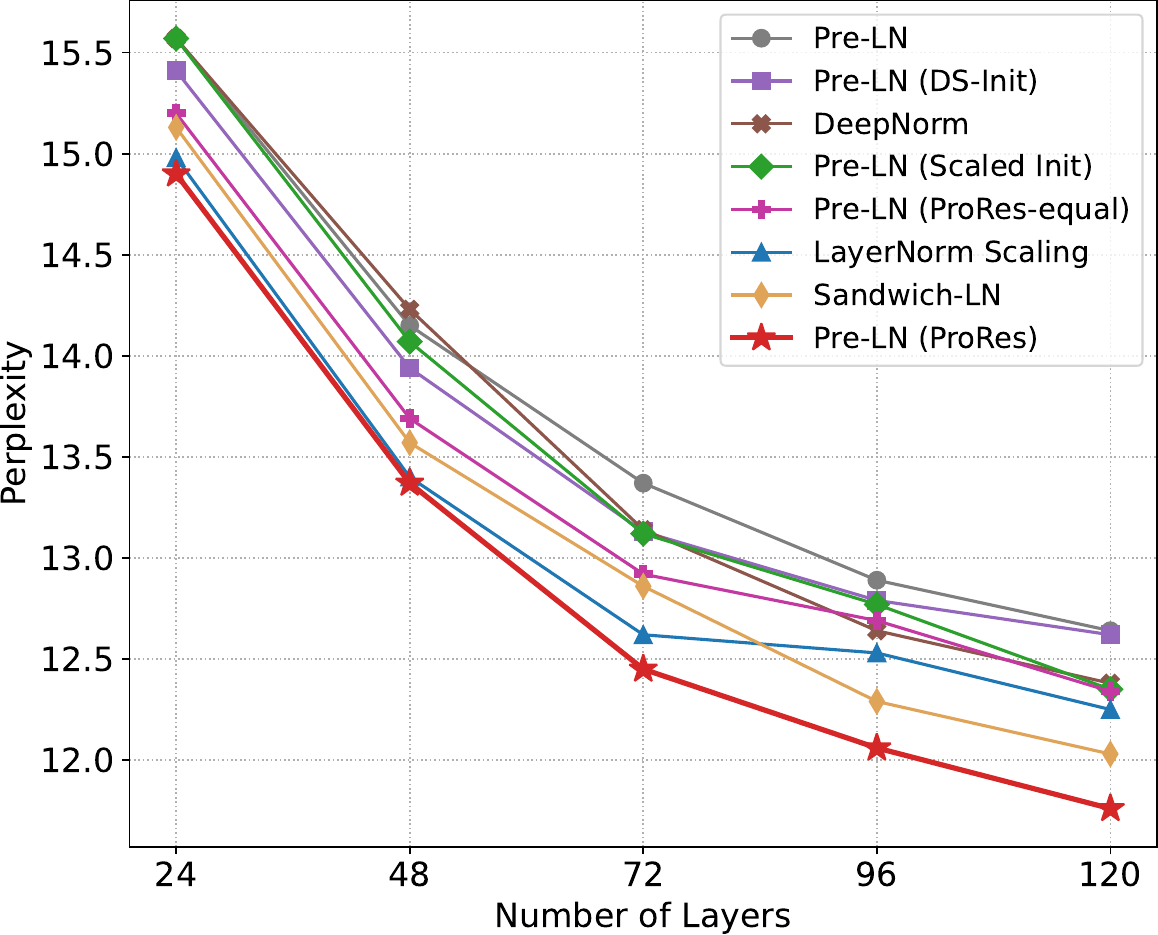}}
    \caption{
      Evaluation perplexity ($\downarrow$) of different methods as model depth increases. See \cref{sec:depth_scaling_exp} for details.
    }
    \label{fig:depth_scaling}
  \end{center}
  \vskip -0.2in
\end{figure}

The Transformer~\cite{vaswani2017attention} has become one of the most dominant backbones for Large Language Models (LLMs), underpinning recent advances across natural language processing and beyond. Among the many components of the Transformer architecture, residual connection~\cite{he2015resnet} and normalization~\cite{ba2016layernorm} play a key role in its optimization, enabling scaling LLMs up to trillion parameters~\cite{kimiteam2025kimik2}.

Despite these advances, scaling Transformers poses unique optimization challenges. A large body of prior work has explored stabilizing Transformer training, including switching from Post-LN to Pre-LN~\cite{xiong2020layernorm}, bounding model updates through depth- or width-aware initializations~\cite{zhang2019rfixup,huang2020tfixup,zhang2019dsinit,liu2020admin,wang2022deepnet,yang2022mup,sun2025lns}, and introducing additional flexibility into residual connections~\cite{wang2019dlcl,bachlechner2020rezero,xie2023residual,zhu2025hyperconnections}. While effective, these methods are largely not \emph{training-phase-aware}, in the sense that their mechanisms are typically applied at initialization, and then leave the optimizer to figure out how to adapt throughout training. In practice, Transformer training exhibits distinct phases~\cite{li2025tracingrepresentationgeometry}. Early training, such as the warmup stage, is often characterized by large and chaotic model updates, whereas the stable training phase involves relatively small and gradual modifications. Moreover, convergence dynamics are heterogeneous across depth, with shallow\footnote{We use ``shallow'' or ``early'', and ``deeper'' or ``later'' interchangeably to describe layer order.} layers tending to converge earlier than deeper layers~\cite{erdogan2025layerlock}. These inherent dynamics suggest that Transformer optimization may benefit from explicitly coordinating layerwise learning across different stages of training.

Motivated by these considerations, we revisit the role of residual connections in layerwise learning from a temporal perspective. Modern LLMs typically consist of sequentially stacked Transformer layers, where each layer updates its input representation\footnote{We use ``representation'', ``hidden state'', and ``activation'' interchangeably. Unless otherwise mentioned, they refer to the token representation at certain Transformer layers.} through residual branches consisting of attention and feed-forward sublayers. While residual connections enable effective optimization at scale, they also allow all layers to modify representations simultaneously from initialization. In the absence of explicit coordination, deeper layers may begin contributing before upstream representations have stabilized, potentially leading to inefficient updates or conflicting learning signals. This observation raises a natural question: can residual contributions be scheduled to respect the staged nature of Transformer training? In particular, can earlier layers be prioritized during early training, while allowing deeper layers to fully engage once upstream representations have matured?

In this work, we propose \textbf{\underline{Pro}}gressive \textbf{\underline{Res}}idual Warmup (\textbf{\method{}}), a simple and scalable approach for coordinating layerwise residual learning over the course of training. \method{} assigns each layer's residual with a predefined scaling factor. This scalar is initialized to 0, and gradually increases to 1 as training progresses. In particular, deeper layers take longer residual warmup than shallow layers, delaying their contribution to representations until shallow layers enter a more stable regime. As a result, shallow layers are prioritized in early training, while deeper layers progressively engage later on. This design controls model updates in the early stages, reduces unnecessary interference across layers, and preserves the expressive capacity of deep layers in later phases of optimization.

We evaluate \method{} through extensive language model pretraining experiments across a range of model scales, initialization methods, and normalization schemes.
As shown in Figure~\ref{fig:depth_scaling}, \method{} consistently improves depth scaling performance, enabling deeper models to achieve better perplexity without compromising training stability.
We further analyze the optimization dynamics introduced by \method{}, shedding light on how coordinating learning order across depth affects learning behavior and representation evolution.

Our main contributions can be summarized as follows:
\begin{itemize}[noitemsep, topsep=0pt]
    \item We propose \method{}, a residual learning scheme that explicitly coordinates layerwise contributions while respecting the staged nature of Transformer convergence.
    \item Extensive pretraining experiments from 71M to 7B parameters show that \method{} improves performance across a wide range of model scales, initialization methods, and normalization schemes.
    \item We analyze the learning dynamics induced by \method{}, providing insight into how learning order across layers influences training stability and efficiency.
\end{itemize}

\section{Related Work}

\paragraph{Transformer initialization and normalization.}
The objective of better initialization or normalization schemes is to stabilize training or accelerate convergence, by controlling the magnitude of activations, weights, gradients and model updates, etc. The original Post-LN~\cite{vaswani2017attention} stabilizes activations by performing normalization after the residual connection. In more recent works, Pre-LN and its variants are adopted because of their stability and efficiency at scale~\cite{radford2019gpt2,xiong2020layernorm,henry2020qknorm,ding2021cogview}. To enable effective depth scaling, several works proposed to scale layerwise weights according to the total number of layers, either uniformly~\cite{radford2019gpt2,shoeybi2020megatronlm,huang2020tfixup,wang2022deepnet}, or layer-specific~\cite{zhang2019dsinit,open_lm}. \citet{yang2022mup} proposed an initialization scheme that enables hyperparameter transfer among various scales. \citet{olmo2025olmo2} found that a simple normal initialization for all weights is more stable and transferable in their setting. A number of works investigated enhancing or replacing normalization with learnable gates or predefined scalings on the residual connection~\cite{bachlechner2020rezero,de2020skipinit,touvron2021layerscale,liu2023branchnorm,sun2025lns,chen2025gpas}. Efforts have been made to combine the advantages of Pre-LN and Post-LN~\cite{xie2023residual,li2025mixln,li2026siamesenorm}. Several works explored enhancing representation capacity by expanding the width of residual stream~\cite{wang2019dlcl,zhu2025hyperconnections,xie2025mhc}.

\paragraph{Progressive growing and freezing layers.}
A similar line of research focuses on progressive training of Transformers. \citet{gong2019stacking} proposed to train deeper BERT models by progressively stacking shallower ones, resulting in an overall shorter training time. \citet{yang2020stacking2} further proposed only training the newly added layers on top for improved efficiency. Similarly, \citet{erdogan2025layerlock} found that shallow layers converge earlier than deeper layers, and proposed a progressive freezing training method to reduce convergence wall time.

\section{Method}

\subsection{Overview}
We begin by describing the implementation of \method{} using Pre-LN Transformer as an example. The forward equation of a Pre-LN layer is:
\begin{equation}
    x_{l+1} = x_l + \mathcal{F}(\mathrm{Norm}(x_l)),
\end{equation}
where $x_l$ is the input representation for layer $l$, $\mathcal F$ denotes the attention or FFN module, and $\mathrm{Norm}$ denotes normalization layer. \method{} modifies the residual connection with a scalar:
\begin{equation}
    x_{l+1} = x_l + \alpha(l,t)\cdot\mathcal{F}(\mathrm{Norm}(x_l)).
\end{equation}
Here $\alpha(l,t)\in\mathbb{R}$ is a predefined scalar that depends on the layer index $l$ and the training step $t$. An example schedule for this variable is:
\begin{equation}\label{eq:sched_linear}
    \alpha(l,t) = \min\!\left(\frac{t}{T\times l},1\right),\quad l=1,\dots,L,
\end{equation}
where $T$ is the warmup length for the first layer and $L$ is the total number of layers. In other words, $\alpha(l,t)$ warms up from 0 to 1 linearly during training, and the warmup length increases linearly with layer index. Under this schedule, residual branches are activated sequentially from shallow to deep layers. We study multiple alternative schedules in \cref{sec:sched_ablation}.

\begin{table}[t]
\caption{Forward equations of Transformer variants, with (\ding{52}) and without (\ding{56}) ProRes.}
\label{tab:eq_forward}
\begin{center}
\resizebox{\linewidth}{!}{%
\begin{tabular}{lcl}
\toprule
Method & ProRes & Forward Equation \\
\midrule
\multirow{2}{*}{Pre-LN} 
  & \ding{56} & $x_{l+1} = x_l + \mathcal{F}(\mathrm{Norm}(x_l))$ \\
  & \ding{52} & $x_{l+1} = x_l + \alpha(l,t)\cdot \mathcal{F}(\mathrm{Norm}(x_l))$ \\
\midrule
\multirow{2}{*}{Post-LN} 
  & \ding{56} & $x_{l+1} = \mathrm{Norm}(x_l + \mathcal{F}(x_l))$ \\
  & \ding{52} & $x_{l+1} = \mathrm{Norm}(x_l + \alpha(l,t)\cdot \mathcal{F}(x_l))$ \\
\midrule
\multirow{2}{*}{Sandwich-LN} 
  & \ding{56} & $x_{l+1} = x_l + \mathrm{Norm}(\mathcal{F}(\mathrm{Norm}(x_l)))$ \\
  & \ding{52} & $x_{l+1} = x_l + \alpha(l,t)\cdot\mathrm{Norm}(\mathcal{F}(\mathrm{Norm}(x_l)))$ \\
\midrule
\multirow{2}{*}{DeepNorm} 
  & \ding{56} & $x_{l+1} = \mathrm{Norm}(\alpha\cdot x_l + \mathcal{F}_\beta(x_l))$ \\
  & \ding{52} & $x_{l+1} = \mathrm{Norm}(\alpha\cdot x_l + \alpha(l,t) \cdot \mathcal{F}_\beta(x_l))$ \\
\midrule
\multirow{2}{*}{LayerNorm Scaling} 
  & \ding{56} & $x_{l+1} = x_l + \mathcal{F}(\mathrm{Norm}(x_l)/\sqrt{l})$ \\
  & \ding{52} & $x_{l+1} = x_l + \alpha(l,t)\cdot \mathcal{F}(\mathrm{Norm}(x_l)/\sqrt{l})$ \\
\bottomrule
\end{tabular}
}%
\end{center}
\end{table}

Applying \method{} to other Transformer variants follows the same principle, as long as they consist of stacked Transformer layers with residual connection. All variants studied in this paper are summarized in \cref{tab:eq_forward}.

\subsection{Motivation of \method{}}\label{sec:motivation}
\method{} is motivated by several principles.

\paragraph{Principle 1: Identity behavior at initialization.}
Deep residual networks benefit from behaving close to identity mapping at initialization, which helps control activation growth and ensures the network has well behaved gradients. Prior work has shown that normalization in residual architectures implicitly biases networks toward identity mappings at initialization~\citep{de2020skipinit}. \method{} makes this bias explicit and exact by initializing the residual scaling parameters as $\alpha(l,t)=0$ when $t=0$, making identity the exact starting point in the network's optimization trajectory.

\paragraph{Principle 2: Bounded model update with respect to depth and time.} 
Stable optimization in deep Transformers requires controlling the overall magnitude of model updates so that training dynamics remain well behaved as model depth increases. Several prior works~\cite{zhang2019rfixup,huang2020tfixup,wang2022deepnet} established that preserving a bounded model update is critical for training deep Transformers. However, these bounds are primarily derived from model behavior at initialization, where updates are typically the most unstable, and may be unnecessarily conservative when applied uniformly throughout training. In practice, Transformer optimization progresses through distinct phases~\cite{li2025tracingrepresentationgeometry}, such as warmup, stable training, and decay~\cite{hu2024minicpm,kosson2024analyzingreducingneed,wen2024understandingwarmupstabledecaylearningrates,dremov2025trainingdynamicscooldownstage}, during which the scale and smoothness of model updates differ substantially. Once the model enters a stable training regime, update constraints designed for initialization may limit learning capacity.\todo{refer to sched ablation} Moreover, empirical studies show that shallow layers tend to converge earlier than deeper layers during training~\citep{erdogan2025layerlock}, which calls for layer-specific treatment. \method{} extends the bounded update principle from initialization to the entire training trajectory in a layerwise and temporal manner. By progressively activating residual branches, \method{} allows shallow layers to stabilize before deeper layers begin to contribute. This encourages representation updates to remain bounded across both depth and training time, stabilizing the warmup phase without sacrificing learning capacity in the stable training regime.

\paragraph{Principle 3: Respecting sequential learning and contribution order.}
In a Transformer with sequentially stacked layers, deeper layers' inputs depend on shallow layers, while shallow layers' gradients depend on deeper layers. Early in training, allowing deeper layers to introduce large residual modifications can inject noise into intermediate representations and skew gradient signals for shallow layers. \method{} enforces a sequential learning order by delaying the contribution of deeper residual branches, ensuring that deeper layers build upon stable representations rather than amplifying early-stage noise with randomly initialized deep residual branches.

\section{Experiments}

We evaluate \method{} across a range of pretraining settings to assess its effectiveness, scalability, and generality. 
First, we conduct pretraining experiments on models ranging from 130M to 1.3B parameters on 50B tokens, comparing and combining \method{} with several Transformer variants. We also scale up training to 7B parameters on Pre-LN in \cref{sec:exp_7b}.
Second, we perform ablations on the residual warmup schedule $\alpha(l, t)$ to study the effect of learning and contribution order across layers.
Third, we examine depth scaling behavior by training models with depths ranging from 12 to 120 layers.
Finally, in \cref{sec:exp_climbmix}, we evaluate \method{} on an alternative pretraining corpus to verify its robustness across data distributions.

\subsection{Experiment Setup}\label{sec:main_exp_setup}

\paragraph{Model architecture.}
All models are decoder-only Transformers with Llama-based Attention and MLP~\cite{grattafiori2024llama3}, which utilize RMSNorm~\cite{zhang2019rmsnorm}, SwiGLU activation~\cite{shazeer2020swiglu} and Rotary Position Embedding~\cite{su2023rope}. The initialization follows~\cite{olmo2025olmo2} where all layers' weights are initialized from truncated normal distribution $\mathcal N(0,0.02^2)$. Depth-dependent initializations are scaled based on this default initialization, including DS-Init~\cite{zhang2019dsinit}, Scaled Init~\cite{shoeybi2020megatronlm}, and DeepNorm~\cite{wang2022deepnet}.

\paragraph{Baselines and \method{} schedule.}
We evaluate \method{} on diverse Transformer variants listed in~\cref{tab:eq_forward}, including Pre-LN~\cite{xiong2020layernorm}, Post-LN~\cite{vaswani2017attention}, Sandwich-LN~\cite{ding2021cogview}, DeepNorm~\cite{wang2022deepnet}, and LayerNorm Scaling (LNS)~\cite{sun2025lns}. For Pre-LN, we additionally experiment with two initialization methods, namely DS-Init (as implemented by~\citet{open_lm}), and Scaled Init. For experiments with \method{}, we adopt the linear schedule in \cref{eq:sched_linear} with $T=1000$. We did not tune this schedule for the main experiment. The effect of schedule choice is studied separately in \cref{sec:sched_ablation}.

\paragraph{Pretraining data.}
We primarily train on the C4-en dataset~\cite{raffel2023t5,dodge2021c4} tokenized with the Llama Tokenizer~\cite{grattafiori2024llama3}. Unless otherwise stated, all experiments train on a randomly sampled subset with 50B tokens and sequences packed to length of 1024. Evaluation perplexities are calculated on a randomly sampled subset of the C4-en test set with 10M tokens. The set of experiments that train on ClimbMix~\cite{diao2025climb} use the same preprocessing protocol, and evaluation perplexities on ClimbMix are calculated on a held out 10M tokens subset.

\paragraph{Optimizer and learning rates.}
All experiments use the AdamW optimizer~\cite{loshchilov2019adamw,kingma2017adam} with $\beta=(0.9,0.95)$, weight decay $0.1$, $\epsilon=10^{-8}$, and gradient clipping $1.0$. A global batch size of 512 with 100k training steps is adopted for all runs. We use the Warmup-Stable-Decay learning rate scheduler~\cite{hu2024minicpm} with 2000 warmup steps and a linear decay to zero in the final $10\%$ steps. Post-LN and DeepNorm use a longer warmup of $10\%$ training steps for better convergence. Learning rates are tuned separately for each configuration with details provided in \cref{sec:lr_tuning}.

\paragraph{Evaluation.}
Other than perplexity measured on test set of the pretraining corpus, we evaluate the pretrained models on general reasoning benchmarks using LM Evaluation Harness~\cite{eval-harness}. We report performance on PIQA~\cite{bisk2019piqa}, SIQA~\cite{sap2019siqa}, HellaSwag~\cite{zellers2019hellaswag}, WinoGrande~\cite{sakaguchi2019winogrande}, ARC-Easy and ARC-Challenge~\cite{clark2018arc}, OpenBookQA~\cite{mihaylov2018obqa}, RACE~\cite{lai2017race}, LAMBADA~\cite{paperno2016lambada}, and MMLU~\cite{hendrycks2021mmlu}. Perplexity is additionally reported on WikiText~\cite{merity2016wikitext} and LAMBADA.

\subsection{Main Results}\label{sec:main_exp}


\definecolor{proresblue}{RGB}{30,144,255}
\newcommand{\shadeProRes}[2]{%
    \pgfmathsetmacro{\shade}{0 + 80*sqrt(min(max(#1,0),1))} 
    \edef\tempcolor{\noexpand\cellcolor{proresblue!\shade}}%
    \tempcolor #2%
}

\begin{table}[t]
  \caption{Perplexity ($\downarrow$) on C4-en test set across model scales. Cells with \method{} are shaded by improvement magnitude.}
  \label{tab:ppl_main}
  \begin{center}
    \begin{small}
      \begin{tabular}{lcccc}
        \toprule
        Method & ProRes & 130M & 350M & 1.3B \\
        \midrule
        \multirow{2}{*}{Pre-LN}
          & \ding{56} & 14.67 & 12.36 & 10.32 \\
          & \ding{52} 
            & \shadeProRes{0.37}{14.30} 
            & \shadeProRes{0.62}{11.74} 
            & \shadeProRes{0.47}{9.86} \\
        \midrule
        \multirow{2}{*}{Pre-LN (DS-Init)}
          & \ding{56} & 14.62 & 12.24 & 10.32 \\
          & \ding{52} 
            & \shadeProRes{0.33}{14.29} 
            & \shadeProRes{0.51}{11.73}
            & \shadeProRes{0.46}{\underline{9.85}} \\
        \midrule
        \multirow{2}{*}{Pre-LN (Scaled Init)}
          & \ding{56} & 14.63 & 12.29 & 10.30 \\
          & \ding{52} 
            & \shadeProRes{0.35}{\underline{14.28}} 
            & \shadeProRes{0.57}{\underline{11.72}} 
            & \shadeProRes{0.46}{\textit{\textbf{\underline{9.84}}}} \\
        \midrule
        \multirow{2}{*}{Sandwich-LN}
          & \ding{56} & 14.55 & 11.97 & 10.16 \\
          & \ding{52} 
            & \shadeProRes{0.05}{14.50} 
            & \shadeProRes{0.19}{11.78} 
            & \shadeProRes{0.22}{9.94} \\
        \midrule
        \multirow{2}{*}{LayerNorm Scaling}
          & \ding{56} & 14.45 & 11.74 &  9.93 \\
          & \ding{52} 
            & \shadeProRes{0.23}{\textit{\textbf{\underline{14.22}}}} 
            & \shadeProRes{0.12}{\textit{\textbf{\underline{11.62}}}}
            & \shadeProRes{0.04}{9.89} \\
        \midrule
        \multirow{2}{*}{Post-LN}
          & \ding{56} & 14.84 & 12.74 & 11.62 \\
          & \ding{52} 
            & \shadeProRes{0.12}{14.72} 
            & \shadeProRes{0.82}{11.92} 
            & \shadeProRes{1.09}{10.53} \\
        \midrule
        \multirow{2}{*}{DeepNorm}
          & \ding{56} & 14.57 & 12.38 & 10.32 \\
          & \ding{52} 
            & \shadeProRes{0.12}{14.45} 
            & \shadeProRes{0.41}{11.97} 
            & \shadeProRes{0.23}{10.09} \\
        \bottomrule
      \end{tabular}
    \end{small}
  \end{center}
  \vskip -0.1in
\end{table}

\begin{table*}[t]
\caption{Zero-shot accuracy ($\uparrow$) on reasoning benchmarks for 1.3B models.}
\label{tab:benchmarks_1b}
\begin{center}
\resizebox{\linewidth}{!}{%
\begin{tabular}{lcccccccccccc}
\toprule
Method & ProRes & PIQA & SIQA & Hella & Wino & ARC-e & ARC-c & OBQA & RACE & Lambada & MMLU & Avg \\
\midrule
\multirow{2}{*}{Pre-LN} 
  & \ding{56} & 72.85 & 41.86 & 52.45 & 55.41 & 53.54 & 26.45 & 33.40 & 31.29 & 39.30 & 27.69 & 43.42 \\
  & \ding{52} & 73.34 & 41.04 & 56.39 & 57.54 & 54.84 & 28.24 & 34.20 & 33.21 & 42.71 & 28.44 & \textbf{\underline{45.00}} \\
\midrule
\multirow{2}{*}{Pre-LN (DS-Init)} 
  & \ding{56} & 72.58 & 41.66 & 52.92 & 56.35 & 53.79 & 27.39 & 34.20 & 32.25 & 39.69 & 27.79 & 43.86 \\
  & \ding{52} & 73.45 & 40.74 & 56.02 & 57.77 & 56.52 & 27.13 & 35.60 & 32.73 & 42.34 & 27.80 & \textit{\textbf{\underline{45.01}}} \\
\midrule
\multirow{2}{*}{Pre-LN (Scaled Init)} 
  & \ding{56} & 72.47 & 39.97 & 53.65 & 54.54 & 52.86 & 26.11 & 32.00 & 32.15 & 37.78 & 27.30 & 42.88 \\
  & \ding{52} & 73.83 & 41.10 & 56.48 & 56.83 & 56.78 & 28.07 & 35.20 & 31.10 & 42.34 & 28.08 & \textbf{44.98} \\
\midrule
\multirow{2}{*}{Sandwich-LN} 
  & \ding{56} & 72.69 & 40.94 & 53.96 & 57.22 & 55.98 & 27.39 & 33.60 & 31.77 & 38.79 & 27.70 & 44.00 \\
  & \ding{52} & 72.31 & 40.99 & 55.55 & 56.12 & 56.06 & 27.65 & 34.60 & 31.87 & 41.63 & 28.17 & \textbf{44.49} \\
\midrule
\multirow{2}{*}{LayerNorm Scaling} 
  & \ding{56} & 73.34 & 41.30 & 55.68 & 56.35 & 55.77 & 27.82 & 33.20 & 33.59 & 38.93 & 28.19 & 44.42 \\
  & \ding{52} & 73.99 & 41.25 & 55.89 & 55.88 & 55.72 & 27.99 & 35.80 & 33.01 & 39.08 & 27.77 & \textbf{44.64} \\
\midrule
\multirow{2}{*}{Post-LN} 
  & \ding{56} & 69.97 & 39.71 & 47.00 & 53.43 & 50.42 & 24.74 & 31.40 & 30.14 & 33.24 & 26.24 & 40.63 \\
  & \ding{52} & 72.74 & 40.94 & 51.81 & 54.62 & 53.58 & 27.56 & 34.40 & 31.77 & 37.82 & 27.56 & \textbf{43.28} \\
\midrule
\multirow{2}{*}{DeepNorm} 
  & \ding{56} & 72.96 & 39.76 & 52.48 & 54.30 & 54.08 & 26.79 & 34.80 & 32.15 & 40.27 & 27.89 & 43.55 \\
  & \ding{52} & 73.23 & 40.43 & 54.71 & 53.99 & 55.89 & 26.37 & 34.80 & 32.73 & 42.31 & 27.98 & \textbf{44.24} \\
\midrule
\rowcolor{proresblue!16}
Average $\Delta$Acc & -- & +0.86 & +0.18 & +2.67 & +0.73 & +1.85 & +0.90 & +1.71 & +0.44 & +2.89 & +0.43 & \textbf{+1.27} \\
\bottomrule
\end{tabular}%
}
\end{center}
\end{table*}

\begin{table}[t]
\caption{Perplexity ($\downarrow$) on WikiText and Lambada for 1.3B models. Final row shows mean perplexity reduction (baseline minus \method{}).}
\label{tab:ppl_wiki_lamb_main}
\centering
\begin{small}
\begin{tabular}{lccc}
\toprule
Method & \method{} & WikiText & Lambada \\
\midrule
\multirow{2}{*}{Pre-LN}
  & \ding{56} & 25.61 & 20.46 \\
  & \ding{52} & \underline{23.67} & \textit{\textbf{\underline{15.05}}} \\
\midrule
\multirow{2}{*}{Pre-LN (DS-Init)}
  & \ding{56} & 25.77 & 19.83 \\
  & \ding{52} & 23.98 & 15.57 \\
\midrule
\multirow{2}{*}{Pre-LN (Scaled Init)}
  & \ding{56} & 25.63 & 20.93 \\
  & \ding{52} & \textit{\textbf{\underline{23.62}}} & \underline{15.42} \\
\midrule
\multirow{2}{*}{Sandwich-LN}
  & \ding{56} & 24.90 & 19.77 \\
  & \ding{52} & 24.12 & 16.09 \\
\midrule
\multirow{2}{*}{LayerNorm Scaling}
  & \ding{56} & 24.28 & 20.79 \\
  & \ding{52} & 24.19 & 18.79 \\
\midrule
\multirow{2}{*}{Post-LN}
  & \ding{56} & 30.37 & 33.53 \\
  & \ding{52} & 26.74 & 22.91 \\
\midrule
\multirow{2}{*}{DeepNorm}
  & \ding{56} & 25.60 & 19.72 \\
  & \ding{52} & 24.77 & 17.15 \\
\midrule
\rowcolor{proresblue!16}
Average $\Delta$PPL & -- & 1.58 & 4.86 \\
\bottomrule
\end{tabular}
\end{small}
\end{table}


For the main experiment, we study the effectiveness of \method{} via pretraining models at scales of 130M, 350M, and 1.3B, and across various Transformer architectures. 

\paragraph{Perplexity on the pretraining corpus.}
\cref{tab:ppl_main} shows perplexity on the test split of the pretraining corpus. Across all configurations, applying \method{} consistently leads to a notable decrease in perplexity, demonstrating its effectiveness as a general residual modification. Among the evaluated architectures, Post-LN benefits the most from \method{}. We attribute this to the normalization in Post-LN naturally biasing contribution towards deeper layers~\cite{xiong2020layernorm,li2025mixln}, whereas \method{} corrects for this inherent tendency via sequentially activating shallow to deep residuals. These improvements are consistent across scales and more effective for larger models. In contrast, combining \method{} with LNS yields smaller gains at larger scales. We hypothesize that since LNS already down-weights deeper residuals by $1/\sqrt{l}$, the additional scaling induced by \method{} further suppresses their contribution, delaying effective learning in deeper layers within the fixed 100k training steps.\todo{remind depth ablation}

\paragraph{Performance on reasoning benchmarks.}
\cref{tab:benchmarks_1b} reports zero-shot accuracy on several reasoning benchmarks. Overall, \method{} provides consistent improvements across architectures, yielding an average gain of 1.27\% over the corresponding baselines. The largest gains are observed on HellaSwag, ARC-Easy, OBQA, and LAMBADA. On LAMBADA, \method{} increases accuracy by 2.89\%, indicating enhanced language modeling of long range dependency. The top three average scores are achieved by Pre-LN models with \method{}, and the top five highest averages are all \method{} variants, highlighting the consistent benefit across architectures.

\paragraph{Generalization to out-of-distribution corpora.}
\cref{tab:ppl_wiki_lamb_main} reports perplexity on WikiText and LAMBADA. While \method{} reduces perplexity moderately on the pretraining corpus ($\Delta\text{PPL}\approx 0.4$), it yields substantially larger gains on these out-of-distribution datasets. Notably, the average perplexity reduction reaches 4.86 on LAMBADA, corresponding well with the observed accuracy improvement on this benchmark.

\subsection{Warmup Schedule Ablation}\label{sec:sched_ablation}

\begin{table}[t]
\caption{ProRes warmup schedules. Here $l \in \{1,\dots,L\}$ denotes the layer index, $t$ the training step, and $T$ the warmup length for the first layer. ``Warmup'' denotes the number of training steps required for all layers to complete residual warmup. See \cref{sec:sched_plot} for visualization.}
\label{tab:eq_schedule}
\centering
\resizebox{\linewidth}{!}{%
\begin{tabular}{lcc}
\toprule
Schedule & Warmup & $\alpha(l,t)$ \\
\midrule
linear
& $T\!\cdot\! L$
& $\min\!\left(\frac{t}{T\times l},1\right)$ \\

linear-sqrt
& $T\!\cdot\! L$
& $\left(\min\!\left(\frac{t}{T\times l},1\right)\right)^{1/2}$ \\

linear-square
& $T\!\cdot\! L$
& $\left(\min\!\left(\frac{t}{T\times l},1\right)\right)^2$ \\

\midrule
equal
& $T$
& $\min\left(t/T,1\right)$ \\

reverse
& $T\!\cdot\! L$
& $\min\!\left(\frac{t}{T\times (L-l+1)},1\right)$ \\

\midrule
stagewise-$0$
& $T\!\cdot\! L$
& $\mathrm{clip}\!\left(\frac{t-T\times (l-1)}{T},0,1\right)$ \\

stagewise-$L$
& $T\!\cdot\! L$
& $\mathrm{clip}\!\left(\frac{t-T\times (l-1)}{T},0,1\right)\!\cdot(1\!-\!1/L)\!+\!1/L$ \\

stagewise-$\sqrt{l}$
& $T\!\cdot\! L$
& $\mathrm{clip}\!\left(\frac{t-T\times (l-1)}{T},0,1\right)\!\cdot(1\!-\!1/\sqrt l)\!+\!1/\sqrt{l}$ \\

\midrule
fix-$L$
& --
& $1/L$ \\

fix-$\sqrt L$
& --
& $1/\sqrt L$ \\

fix-$\sqrt{l}$
& --
& $1/\sqrt{l}$ \\

\bottomrule
\end{tabular}
}%
\end{table}

\definecolor{proresblue}{RGB}{30,144,255}
\definecolor{lightred}{RGB}{220,40,40}

\definecolor{lightgray}{RGB}{220,220,220}

\newcommand{\shadeImprove}[2]{%
  \pgfmathsetmacro{\shade}{80*min(max(#1,0),1)}%
  \edef\temp{\noexpand\cellcolor{proresblue!\shade}}%
  \temp #2%
}

\newcommand{\shadeDegrade}[2]{%
  \pgfmathsetmacro{\shade}{80*sqrt(min(max(#1,0),1))}%
  \edef\temp{\noexpand\cellcolor{lightred!\shade}}%
  \temp #2%
}

\newcommand{\shadeDiverged}{\cellcolor{lightgray}{\textit{diverged}}}

\begin{table}[t]
\caption{Ablation of ProRes warmup schedules. Perplexity ($\downarrow$) on C4 for 350M models trained with 6B tokens. ``Warmup'' denotes the number of training steps required for all layers to complete residual warmup. \textcolor{proresblue}{Blue} indicates improvement while \textcolor{lightred}{red} indicates degrading performance relative to no-\method{} baseline. Perplexities higher than 1000 are marked as \textit{diverged}. \textbf{Additional baselines:} DeepNorm (15.71) and LNS (14.63), trained under the same configuration without \method{}.}
\label{tab:ppl_sched_ablation}
\centering
\resizebox{\linewidth}{!}{%
\begin{tabular}{lcccc}
\toprule
Schedule & Warmup & Pre-LN & Post-LN & Sandwich-LN \\
\midrule
--       & --   & 15.21 & 16.83 & 14.76 \\
\midrule
linear   & 1k   & \shadeImprove{0.61}{14.60} & \shadeImprove{0.85/2}{15.98} & \shadeImprove{0.13*1.5}{14.63} \\
linear   & 3k   & \shadeImprove{0.72}{14.49} & \shadeImprove{1.18/2}{15.65} & \shadeImprove{0.20*1.5}{14.56} \\
linear   & 6k   & \shadeImprove{0.77}{14.44} & \shadeImprove{1.50/2}{15.33} & \shadeImprove{0.25*1.5}{14.51} \\
linear   & 12k  & \shadeImprove{0.80}{\textit{\textbf{\underline{14.41}}}} & \shadeImprove{1.68/2}{15.15} & \shadeImprove{0.31*1.5}{\underline{14.45}} \\
linear   & 24k  & \shadeImprove{0.78}{\underline{14.43}} & \shadeImprove{1.71/2}{15.12} & \shadeImprove{0.30*1.5}{14.46} \\
linear   & 48k  & \shadeImprove{0.64}{14.57} & \shadeImprove{1.65/2}{15.18} & \shadeImprove{0.32*1.5}{\textit{\textbf{\underline{14.44}}}} \\
\midrule
linear-sqrt  &  24k  & \shadeImprove{0.57}{14.64} & \shadeImprove{1.39/2}{15.44} & \shadeImprove{0.17*1.5}{14.59} \\
linear-square  &  24k  & \shadeImprove{0.75}{14.46} & \shadeImprove{1.99/2}{\textit{\textbf{\underline{14.84}}}} & \shadeImprove{0.29*1.5}{14.47} \\
\midrule
equal    & 1k   & \shadeImprove{0.42}{14.79} & \shadeImprove{0.54/2}{16.29} & 14.76 \\
equal    & 12k  & \shadeImprove{0.15}{15.06} & \shadeDiverged & \shadeDegrade{0.36}{15.12} \\
equal    & 24k  & \shadeDegrade{0.08}{15.29} & \shadeDiverged & \shadeDegrade{0.38}{15.14} \\
\midrule
reverse  & 1k   & \shadeImprove{0.23}{14.98} & \shadeDiverged & \shadeDegrade{0.20}{14.96} \\
reverse  & 12k  & \shadeDegrade{0.36}{15.57} & \shadeDiverged & \shadeDegrade{1.03/2}{15.79} \\
reverse  & 24k  & \shadeDegrade{0.44}{15.65} & \shadeDiverged & \shadeDegrade{2.02/2}{16.78} \\
\midrule
stagewise-$0$  & 24k   & \shadeDegrade{4.3}{19.51} & \shadeDiverged & \shadeDegrade{3.55}{18.31} \\
stagewise-$L$  & 24k   & \shadeDegrade{0.14}{15.35} & \shadeImprove{1.74/2}{\underline{15.09}} & \shadeDegrade{0.05}{14.81} \\
stagewise-$\sqrt{l}$  & 24k   & \shadeImprove{0.41}{14.80} & \shadeImprove{1.35/2}{15.48} & \shadeImprove{0.05*1.5}{14.71} \\
\midrule
fix-$L$  & --   & \shadeDegrade{0.56}{15.77} & \shadeDegrade{3.35/2}{20.18} & \shadeDegrade{0.71}{15.47} \\
fix-$\sqrt L$  & --   & \shadeDegrade{0.27}{15.48} & \shadeDegrade{0.21/2}{17.04} & \shadeDegrade{0.23}{14.99} \\
fix-$\sqrt l$  & --   & \shadeImprove{0.23}{14.98} & \shadeImprove{0.62/2}{16.21} & \shadeDegrade{0.06}{14.82} \\
\bottomrule
\end{tabular}
}%
\end{table}

We now study the influence of different \method{} schedules $\alpha(l,t)$ on training dynamics and model performance, with the goal of empirically validating the design motivations discussed in \cref{sec:motivation}.
We conduct ablation on a diverse set of schedules listed in \cref{tab:eq_schedule}, and across 3 representative Transformer variants: Pre-LN, Post-LN, and Sandwich-LN. For all ablation runs, we pretrain 350M models with 24 layers ($L=24$) on 6B tokens over 60k training steps, which aligns with the Chinchilla Scaling Law~\cite{hoffmann2022chinchilla}. Perplexities on test set are reported in \cref{tab:ppl_sched_ablation}. We summarize the main observations regarding layer activation order, warmup length, and schedule design below.

\paragraph{Effect of layer contribution order.}
We observe that the order in which residual branches are activated across depth plays a significant role in optimization. Schedules that activate residuals progressively from shallow to deep layers (``linear'') consistently outperform schedules that activate all layers simultaneously\footnote{We note that BranchNorm~\cite{liu2023branchnorm} can be viewed as a special case of Post-LN with an ``equal'' \method{} schedule.} (``equal'') or prioritize deeper layers early in training (``reverse''). This trend holds across all three architectures, indicating that respecting the sequential dependency structure of stacked Transformer layers is beneficial for learning.

\paragraph{Warmup length and interaction with layer order.}
The ``linear'' schedule remains effective across a wide range of warmup lengths, provided the warmup is neither too short ($<1$k steps) nor so long that it occupies a substantial fraction of training ($>48$k steps). In contrast, the ``equal'' schedule is considerably more sensitive to warmup length and can lead to degraded performance or even divergence as warmup increases. We conjecture that the ``equal'' schedule primarily delays chaotic updates at initialization, but fails to respect the sequential dependency across layers; as a result, simultaneously scaling up all residuals may amplify representation and gradient noise. The ``reverse'' schedule, which activates deeper layers prior to shallow ones, becomes increasingly detrimental as warmup length increases. A longer warmup leads to a clearer separation of learning phases across layers, allowing deeper layers to dominate optimization early in training. Once this imbalance is established, it becomes difficult to correct by reactivating shallow layers later. This behavior mirrors a typical divergence pattern observed in Post-LN Transformers~\cite{xiong2020layernorm,wang2022deepnet}, where the last few layers dominate model update, leaving shallow layers with vanishing gradients and limited signal to escape poor local minima.

\paragraph{Static vs. dynamic constraint on model update.}
In Principle 2 of \cref{sec:motivation}, we hypothesized that techniques designed to bound model updates at initialization may be unnecessarily conservative once training enters a stable regime. We empirically verify this by observing that the ``fix'' schedules always perform worse than their ``stagewise'' counterparts (e.g. ``fix-$\sqrt l$'' and ``stagewise-$\sqrt l$''), especially for Post-LN.  The ``fix'' schedules constrain model update by scaling down the residual outputs according to model depth throughout the entire training process. ``stagewise'' variants apply an equivalent constraint at initialization, but progressively relax it by restoring residual magnitudes from shallow to deep layers during training. This dynamic treatment stabilizes the warmup phase comparably to ``fix'' schedules, without hurting learning potential in the stable training phase. Moreover, DeepNorm provides a more principled static approach for controlling model updates of Post-LN, yet still underperforms several \method{} schedules, further suggesting that static methods might fail to fully utilize the learning potential in the stable training phase.

\paragraph{Identifying the optimal \method{} schedule.}
The ``linear'' schedule is the most robust overall, achieving strong performance across all three architectures and a wide range of warmup lengths, and is the best-performing evaluated schedule for Pre-LN and Sandwich-LN. For Post-LN, ``linear-square'' and ``stagewise-$L$'' perform notably better than ``linear'', highlighting the importance of a sufficiently gentle introduction of residuals during early training, which corresponds to Principle~1 in \cref{sec:motivation}. Overall, the optimal schedule is architecture-dependent, with ``linear'' serving as a strong default choice.

\subsection{Depth Scaling Experiment}\label{sec:depth_scaling_exp}

\begin{figure}[ht]
  \vskip 0.2in
  \begin{center}
    \centerline{\includegraphics[width=\columnwidth]{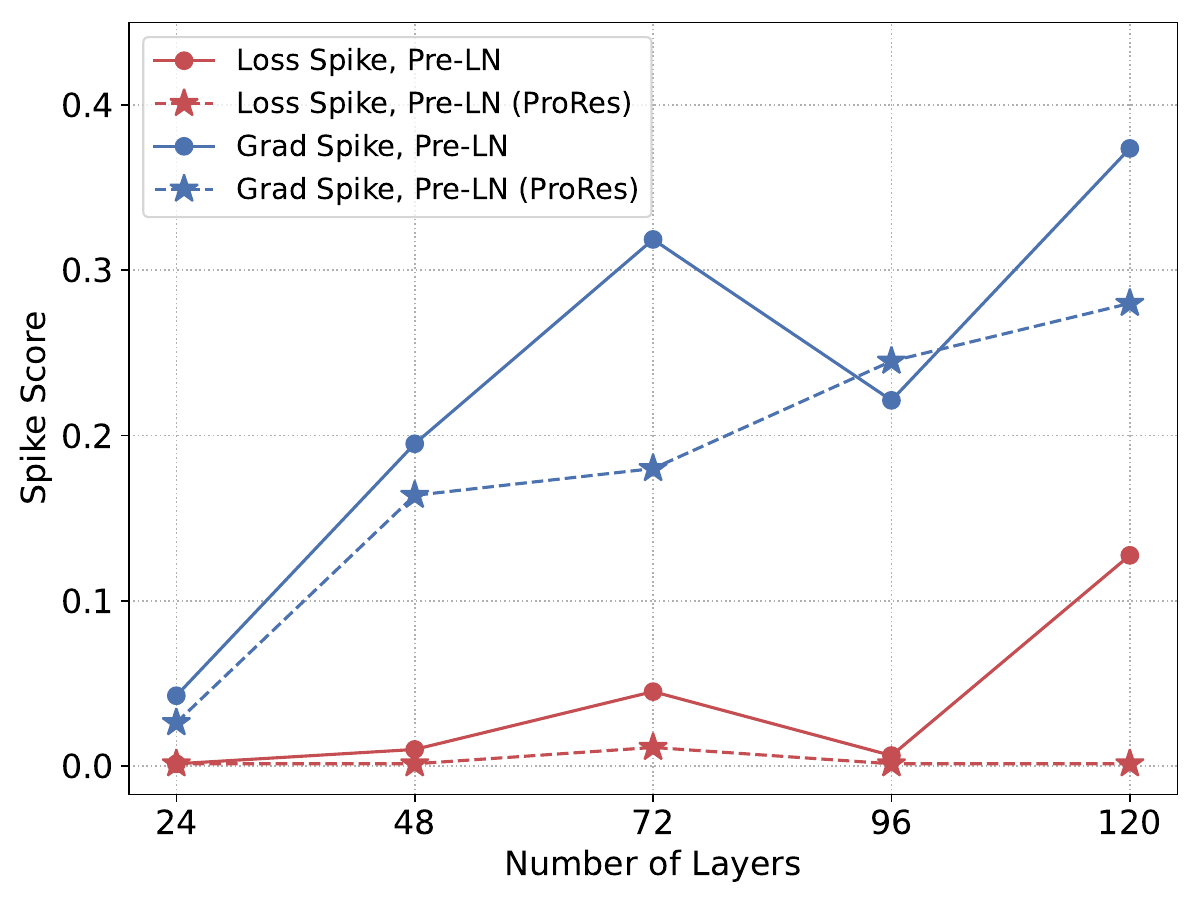}}
    \caption{
      Spike score ($\downarrow$) of loss and gradient norm as model depth increases.
    }
    \label{fig:spike_score}
  \end{center}
\end{figure}

We investigate how \method{} facilitates residual learning as model depth increases, by comparing its depth scaling behavior against several baseline methods. Results are summarized in \cref{fig:depth_scaling}.

\paragraph{Setup.}
We start from a base configuration of a 71M-parameter Transformer with 12 layers and scale directly to 24, 48, 72, 96, and 120 layers, keeping all other dimensions fixed. For visual clarity in \cref{fig:depth_scaling}, \method{} is applied only to Pre-LN, the best-performing combination according to \cref{sec:main_exp}, while studies across other normalization schemes are left to future work. Two \method{} schedules are evaluated: ``linear'' (\textit{Pre-LN (\method{})}) and ``equal'' (\textit{Pre-LN (\method{}-equal)}). The same set of no-\method{} baselines from \cref{sec:main_exp_setup} is adopted, excluding Post-LN.  
All models are trained under the same setting as \cref{sec:main_exp_setup}, with a learning rate of $2\times10^{-3}$ for all depths except DeepNorm, which required additional search from $\{5\times10^{-4},1\times 10^{-3},1.5\times 10^{-3},2\times 10^{-3}\}$ for optimal convergence.  
For \textit{Pre-LN (\method{})}, we use the ``linear'' schedule with $T=1000$ at depths up to 72, and $T=500$ for 96 and 120 layers to ensure the warmup length does not exceed the total training steps. No additional tuning is performed for the ``linear'' schedule. For \textit{Pre-LN (\method{}-equal)}, we found that $T=1000$ leads to degraded performance at 120 layers, and instead used $T=12$k for all depths.

\paragraph{Depth scaling performance.}
\cref{fig:depth_scaling} shows that \textit{Pre-LN (\method{})} consistently delivers the best performance among all methods, with the lead becoming more pronounced as depth increases. LNS matches \method{} at depths up to 48 layers, but falls short beyond 72 layers. We hypothesize that the static scaling of LNS, while controlling activation growth at initialization, overly downscales deeper residuals as model depth increases, limiting their contribution and learning potential. This is also observed in the scaling trend for \textit{Pre-LN (DS-Init)}, which initializes deeper layer weights to smaller values, leading to slower perplexity reduction beyond 72 layers. Sandwich-LN enforces a constant norm on residual outputs and provides better scaling potential than LNS at larger depths. Depth-dependent initializations improve the performance of Pre-LN mildly. DeepNorm surpasses Pre-LN from 72 layers onwards but requires careful tuning of learning rates for convergence.

\paragraph{Training stability.}
To understand how \method{} affects training stability, we measure loss and gradient spikes using the spike score defined by \citet{olmo2025olmo2}, which quantifies the percentage of data points that are at least seven standard deviations away from a rolling average of the previous 1000 steps. We compute spike scores over the stable training phase (10\%--90\% of training progress). As shown in \cref{fig:spike_score}, applying \method{} maintains near-zero loss spikes as depth increases, indicating that \method{} not only enhances performance but also stabilizes training.

\section{Analysis}

In this section, we analyze the training dynamics that \method{} introduces. First, we show that applying \method{} naturally mitigates the exponential activation growth often observed in Pre-LN. Second, we study the evolution of layerwise representations during training, and show that \method{} enables smoother and more stable representation updates.

\begin{figure*}[ht]
  \vskip 0.2in
  \begin{center}
    \begin{minipage}{0.48\textwidth}
      \includegraphics[width=\textwidth]{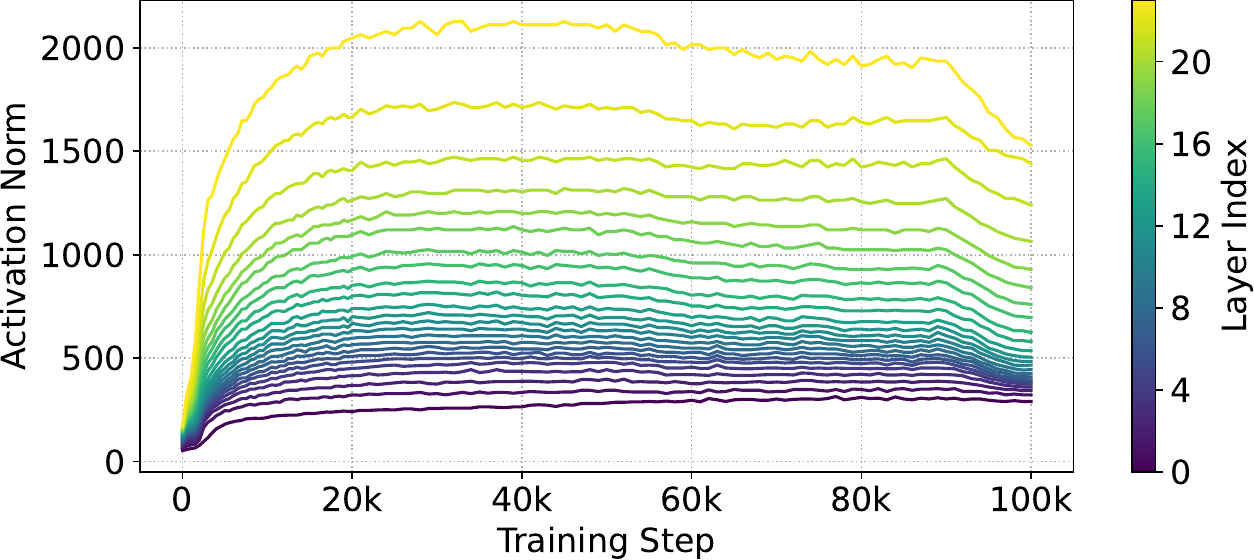}
      \subcaption{Pre-LN without \method{}}
      \label{fig:act_norm_pre}
    \end{minipage}
    \hfill
    \begin{minipage}{0.48\textwidth}
      \includegraphics[width=\textwidth]{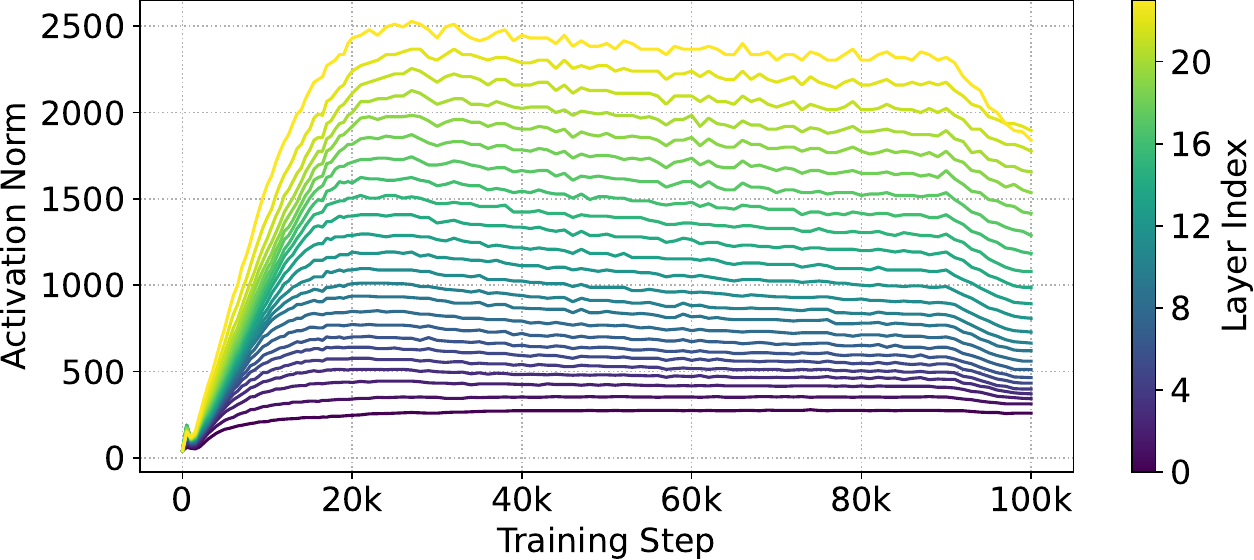}
      \subcaption{Pre-LN with \method{}}
      \label{fig:act_norm_pre_prores}
    \end{minipage}
    \caption{
      Layerwise activation norm of Pre-LN, with and without \method{}.
    }
    \label{fig:act_norm}
  \end{center}
\end{figure*}

\begin{figure*}[ht]
  \vskip 0.2in
  \begin{center}
    \begin{minipage}{0.48\textwidth}
      \includegraphics[width=\textwidth]{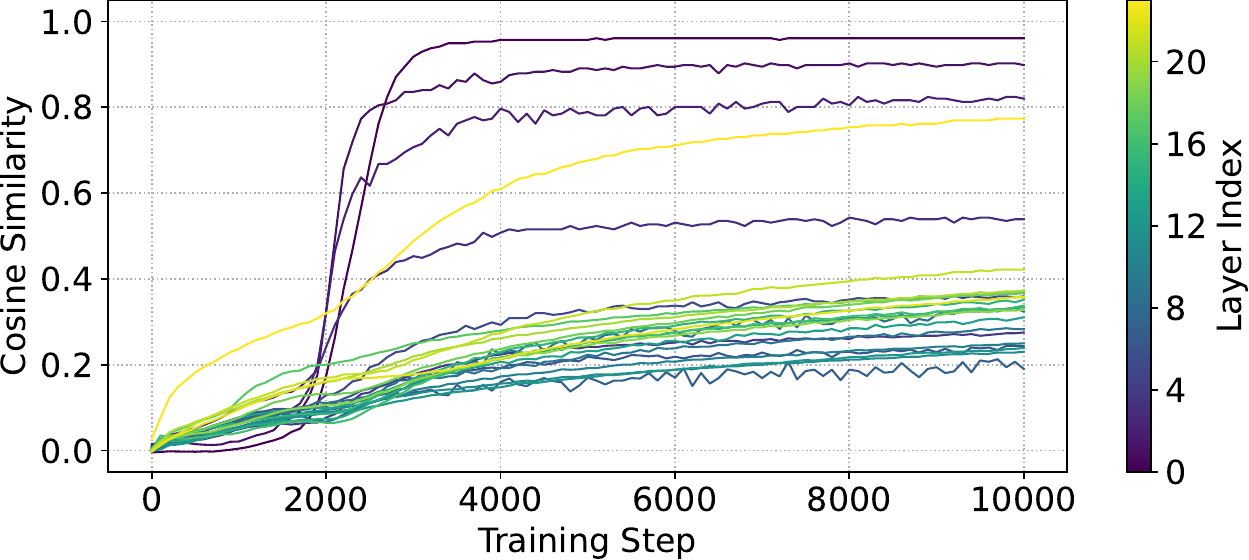}
      \subcaption{Pre-LN without \method{}}
      \label{fig:repr_evo_pre}
    \end{minipage}
    \hfill
    \begin{minipage}{0.48\textwidth}
      \includegraphics[width=\textwidth]{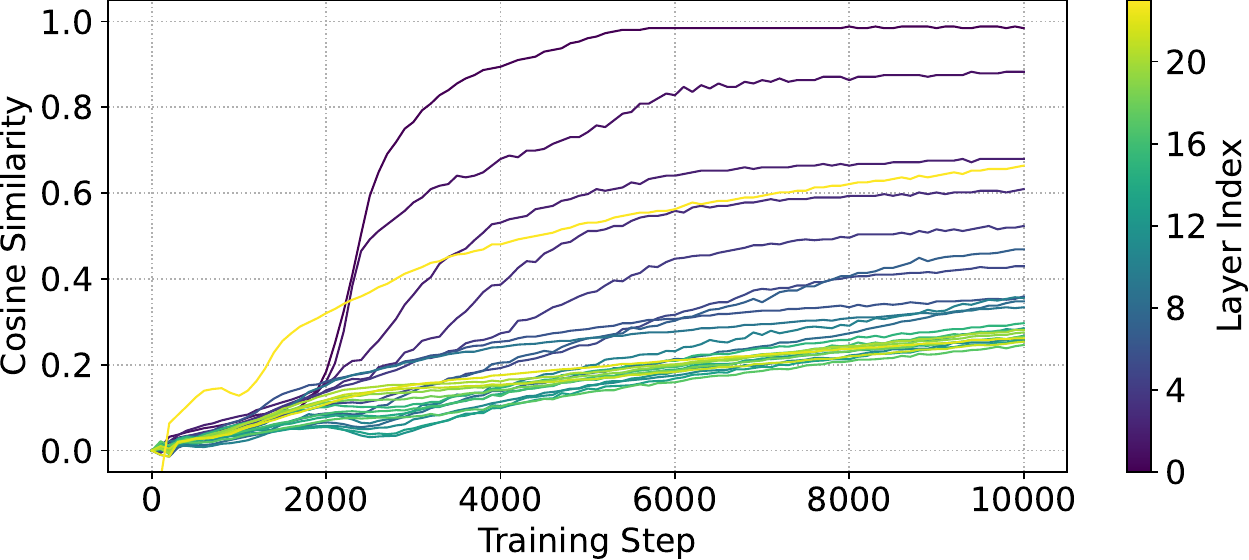}
      \subcaption{Pre-LN with \method{}}
      \label{fig:repr_evo_pre_prores}
    \end{minipage}
    \caption{
      Cosine similarity of layerwise residuals between intermediate (first 10\% steps) and final checkpoint.
    }
    \label{fig:repr_evo}
  \end{center}
\end{figure*}

\subsection{Activation Growth in Pre-LN}

Prior works have shown that Pre-LN Transformers often exhibit exponential activation growth across depth, rather than a desired linear growth~\cite{li2025mixln,sun2025lns}, which can limit the effective contribution of deeper layers~\cite{gromov2025unreasonable,men2024shortgpt}. 

We demonstrate that applying \method{} naturally resolves this issue. \cref{fig:act_norm} shows the layerwise activation norms during training for 1.3B models in \cref{sec:main_exp}. In early training (0--20k steps), activation norms of vanilla Pre-LN quickly grows exponentially, while the \method{} variant facilitates a gentle and more linear growth. We attribute this behavior to differences in how residual updates are coordinated across depth. In vanilla Pre-LN, all layers update representations concurrently from initialization, even when upstream representations are still highly unstable. As a result, deeper layers may apply updates based on rapidly changing inputs, while shallow layers are simultaneously influenced by fluctuating gradient signals from deeper layers. By contrast, \method{} delays the contribution of deeper residual branches during early training, allowing shallow layers to first stabilize representations before deeper layers begin refining them.
This staged interaction leads to more controlled activation growth across depth.


\subsection{Evolution of Layerwise Representations}

To further examine how \method{} affects training dynamics, we analyze the evolution of layerwise residual outputs over the course of training.
Specifically, we measure the cosine similarity between residual outputs at intermediate checkpoints and those of the final fully trained model.
\cref{fig:repr_evo} shows that both vanilla Pre-LN and \method{} exhibit earlier convergence in shallow layers.
However, the representation evolution under \method{} is notably smoother across layers.
In contrast, vanilla Pre-LN shows frequent fluctuations in representation similarity throughout training, indicating less efficient evolution of residual outputs.
These observations empirically support our hypothesis that explicitly coordinating residual contributions across depth reduces counterproductive updates and leads to more stable representation learning.

















\section{Conclusions}

We introduced \method{}, a simple and scalable residual warmup scheme that explicitly coordinates layerwise learning over the course of Transformer training. By progressively activating residual contributions across depth, \method{} prioritizes shallow layers early while allowing deeper layers to engage once upstream representations stabilize. Extensive pretraining experiments show that \method{} consistently improves performance and depth scaling across model sizes, initialization methods, and normalization schemes. These results suggest that training-phase-aware residual scheduling is an effective and practical direction for improving Transformer optimization.

\section*{Acknowledgements}
This work was partially supported by an Area of Excellence project (AoE/E-601/24-N), a Theme-based Research Project (T32-615/24-R) and the Innovation and Technology Commission (ITCPD/17-9) from the Research Grants Council of the Hong Kong Special Administrative Region, China.



\section*{Impact Statement}

This paper presents a method for improving the optimization and depth scaling of Transformer-based language models. The primary goal of this work is to advance the understanding and design of scalable training techniques in machine learning. While improved training stability and efficiency may facilitate the development of larger and more capable models, the broader societal impacts are consistent with those of existing large language models. We do not identify any new ethical concerns or societal risks that are unique to the techniques proposed in this work.





\bibliography{references}
\bibliographystyle{icml2026}

\newpage
\appendix
\onecolumn

\section{Learning Rate Tuning}\label[appendix]{sec:lr_tuning}
In this section, we detail the learning rate tuning for the main experiments in \cref{sec:main_exp_setup}. We first note that the final performance of the WSD learning rate scheduler~\cite{hu2024minicpm} is not quite sensitive to learning rate choices. A smaller learning rate tends to have a lower loss during the constant learning rate phase, but a larger learning rate would catch up after the decay stage, as long as it trains stably. A similar phenomenon is discussed in~\cite{olmo2025olmo2}. With this property in mind, our goal is to find a preferably larger learning rate such that training takes place smoothly, without notable loss spikes or gradient spikes. We only tune the learning rates for no-\method{} baselines, and directly apply their tuned learning rates to \method{} variants.

For the 1.3B models, we searched over learning rates $\{1\times 10^{-3}, 6\times 10^{-4}, 3\times 10^{-4}\}$. We selected $6\times 10^{-4}$ for all baselines except Post-LN and DeepNorm, for which $3\times 10^{-4}$ yielded more stable training.

For the 350M models, we evaluated $\{5\times 10^{-4}, 1\times 10^{-3}, 1.5\times 10^{-3}\}$ and adopted $1\times 10^{-3}$ for all models except Post-LN, which performed best with $5\times 10^{-4}$.

For the 130M models, we considered $\{1\times 10^{-3}, 1.5\times 10^{-3}, 2\times 10^{-3}\}$. The final choices were $1\times 10^{-3}$ for Post-LN, $1.5\times 10^{-3}$ for DeepNorm, and $2\times 10^{-3}$ for the remaining baselines.

\section{Pretrain Experiments on 7B parameters}\label[appendix]{sec:exp_7b}

\begin{figure*}[ht]
  \vskip 0.2in
  \begin{center}
    \begin{minipage}{0.48\textwidth}
      \includegraphics[width=\textwidth]{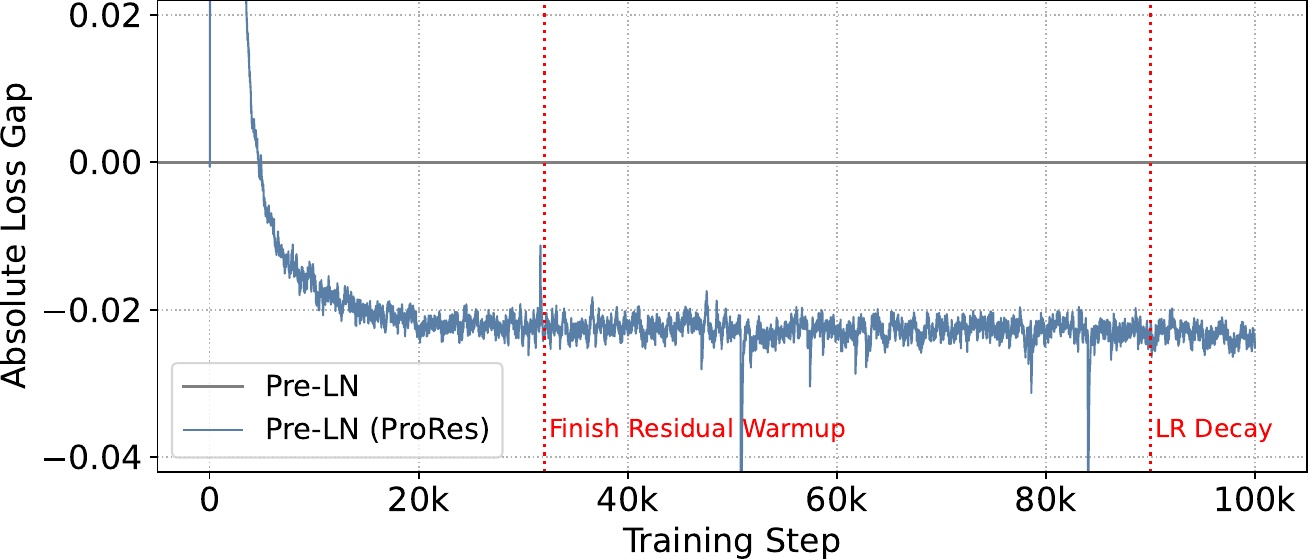}
    \end{minipage}
    \hfill
    \begin{minipage}{0.48\textwidth}
      \includegraphics[width=\textwidth]{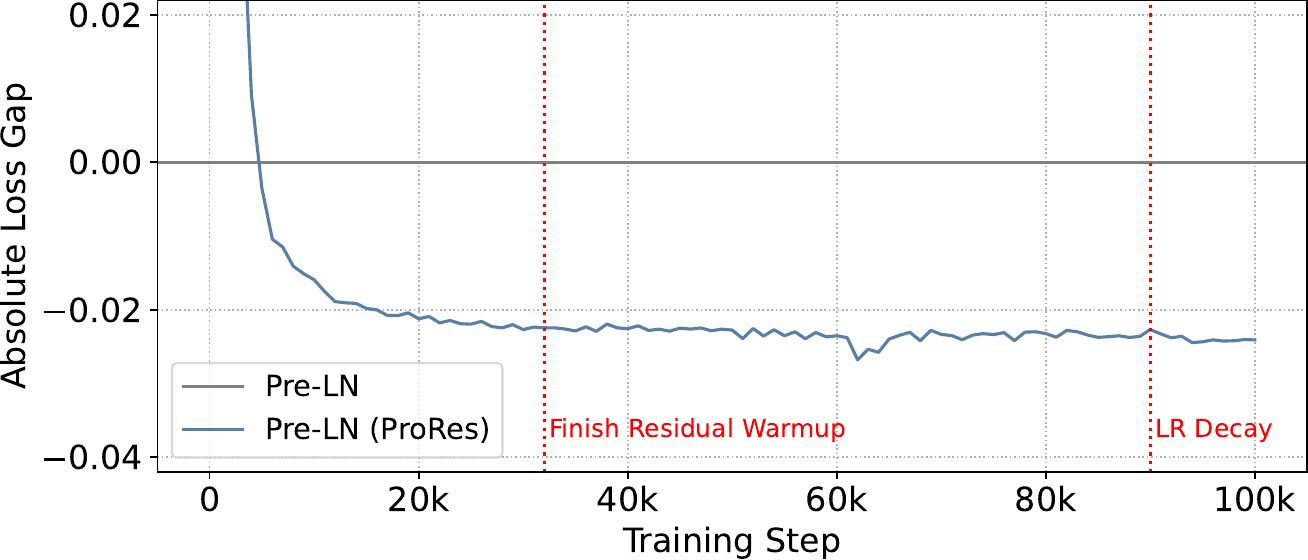}
    \end{minipage}
    \caption{
      Loss gap across training steps of 7B models. \textbf{Left:} training loss gap (0.99 EMA smoothed) between Pre-LN and its \method{} variant. \textbf{Right:} evaluation loss gap calculated every 1000 steps.
    }
    \label{fig:loss_gap_7b}
  \end{center}
\end{figure*}

We conduct pretraining experiments on 7B parameter models following the settings in \cref{sec:main_exp_setup}, using a learning rate of $3\times10^{-4}$. Vanilla Pre-LN and Pre-LN with \method{} are evaluated. For the \method{} variant, we adopt the linear schedule in \cref{eq:sched_linear} with $T=1000$, which corresponds to a residual warmup length of 32,000 steps given 32 layers in total.

\cref{fig:loss_gap_7b} shows the loss gaps during training. We note that \method{} has higher loss initially due to its restricted model updates when $\alpha(l,t)$ is small. As training proceeds, \method{} gradually unlocks learning potentials of each layer and achieves lower loss than the baseline. The loss gap keeps increasing steadily even after all residuals have finished warming up. Moreover, \method{} is able to maintain and even slightly increase its advantage after the learning rate decay stage.

\section{Pretrain Experiments on Alternative Corpus}\label[appendix]{sec:exp_climbmix}

\begin{table}[t]
  \caption{Perplexity ($\downarrow$) on ClimbMix test split across model scales.}
  \label{tab:ppl_climb}
  \begin{center}
    \begin{small}
      \begin{tabular}{lcccc}
        \toprule
        Method & ProRes & 130M & 350M & 1.3B \\
        \midrule
        \multirow{2}{*}{Pre-LN}
          & \ding{56} & 11.11 & 9.43 & 8.38 \\
          & \ding{52} & 10.82 & 9.13 & 8.09 \\
        \bottomrule
      \end{tabular}
    \end{small}
  \end{center}
  \vskip -0.1in
\end{table}

To verify that \method{} generalizes well to pretraining corpus other than C4, we conduct a set of experiments on the ClimbMix dataset~\cite{diao2025climb}. We adopt the same preprocessing pipeline in \cref{sec:main_exp_setup}. Evaluation perplexities on held out test set are reported in \cref{tab:ppl_climb}.

\section{Visualization of \method{} Schedules}\label[appendix]{sec:sched_plot}

\begin{figure}[ht]
  \vskip 0.2in
  \begin{center}
    \centerline{\includegraphics[width=\columnwidth]{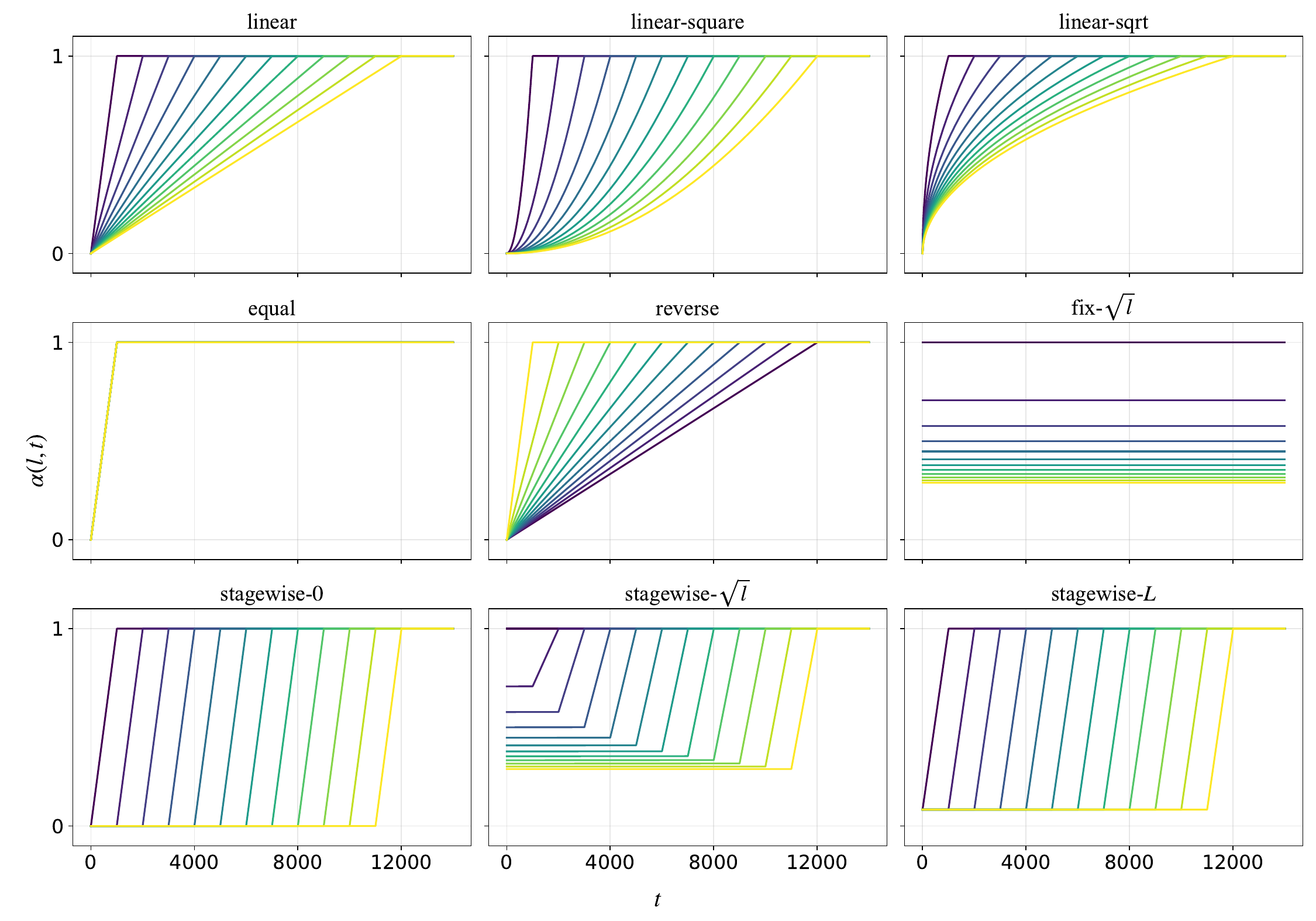}}
    \caption{
      Visualization of various schedules in \cref{tab:eq_schedule}. We use $T=1000$, $L=12$ for visualizing $\alpha(l,t)$. Darker lines indicate shallow layers while brighter lines indicate deeper layers.
    }
    \label{fig:sched}
  \end{center}
\end{figure}

\section{Use of Compute Resources}
All experiments were conducted on a single node of 8 $\!\times\!$ H800 GPUs. Training under the settings in \cref{sec:main_exp_setup} required approximately 260h for 7B, 100h for 3B, 50h for 1.3B, 20h for 350M, and 7h for 130M models.



\end{document}